\documentclass[10pt,twocolumn,letterpaper]{article}

\usepackage[accsupp]{axessibility}

\usepackage{iccv}
\usepackage{times}
\usepackage{epsfig}
\usepackage{graphicx}
\usepackage{amsmath}
\usepackage{amssymb}

\usepackage{multirow}
\usepackage{siunitx,booktabs,caption}

\usepackage{setspace}
\usepackage{ctable}
\usepackage{makecell}
\usepackage{caption,subcaption}

\usepackage[pagebackref,breaklinks,colorlinks]{hyperref}  

\iccvfinalcopy 

\def\iccvPaperID{5860} 

\ificcvfinal\fi

\begin{document}

\title{GAFlow: Incorporating Gaussian Attention into Optical Flow}

\author{Ao Luo\textsuperscript{\rm 1}, Fan Yang\textsuperscript{\rm 2}, Xin Li\textsuperscript{\rm 2}, Lang Nie\textsuperscript{\rm 3}, Chunyu Lin\textsuperscript{\rm 3}, Haoqiang Fan\textsuperscript{\rm 1}, and Shuaicheng Liu\textsuperscript{\rm 4,1}\thanks{Corresponding author} \\
	\textsuperscript{\rm 1}Megvii Technology \qquad \textsuperscript{\rm 2}Group 42 \qquad
	\textsuperscript{\rm 3}Beijing Jiaotong University \\
	\textsuperscript{\rm 4}University of Electronic Science and Technology of China\\
}

\maketitle
\ificcvfinal\thispagestyle{empty}\fi

\begin{abstract}
   Optical flow, or the estimation of motion fields from image sequences, is one of the fundamental problems in computer vision. Unlike most pixel-wise tasks that aim at achieving consistent representations of the same category, optical flow raises extra demands for obtaining local discrimination and smoothness, which yet is not fully explored by existing approaches. In this paper, we push Gaussian Attention (GA) into the optical flow models to accentuate local properties during representation learning and enforce the motion affinity during matching. Specifically, we introduce a novel Gaussian-Constrained Layer (GCL) which can be easily plugged into existing Transformer blocks to highlight the local neighborhood that contains fine-grained structural information. Moreover, for reliable motion analysis, we provide a new Gaussian-Guided Attention Module (GGAM) which not only inherits properties from Gaussian distribution to instinctively revolve around the neighbor fields of each point but also is empowered to put the emphasis on contextually related regions during matching. Our fully-equipped model, namely Gaussian Attention Flow network (GAFlow), naturally incorporates a series of novel Gaussian-based modules into the conventional optical flow framework for reliable motion analysis. 
   Extensive experiments on standard optical flow datasets consistently demonstrate the exceptional performance of the proposed approach in terms of both generalization ability evaluation and online benchmark testing. 
   Code is available at \url{https://github.com/LA30/GAFlow}.
\end{abstract}

\section{Introduction}

\label{sec:intro}

Optical flow aims to establish pixel-wise correspondences across images, playing a crucial role in video understanding. It unifies representation learning and feature matching as a problem of pixel-wise motion inference. Modern optical flow models typically focus on either improving representation learning techniques (\eg, alternative learning~\cite{Sun2018PWCNetCF,Hur2019IterativeRR,Zhao2020MaskFlownetAF} and reinforcement learning~\cite{amiranashvili2018motion}) or refining feature similarity measurement methodologies (\eg, 4D correlation volumes~\cite{Teed2020RAFTRA} or 4D Transformer~\cite{huang2022flowformer}). Despite these significant advancements, a glaring limitation persists: these models largely neglect the exploration of local structural information. This oversight hinders their performance, particularly in challenging scenarios involving large motions, occlusions, blurring effects, and shifts in appearance. Such situations demand an elevated focus on local discrimination and flow consistency. This brings us to an intriguing inquiry: \emph{Is it feasible to architect optical flow models that intrinsically focus on local structural information during both representation learning and feature matching?}

\begin{figure}[pt]
	\begin{center}
		\includegraphics[width=1\linewidth]{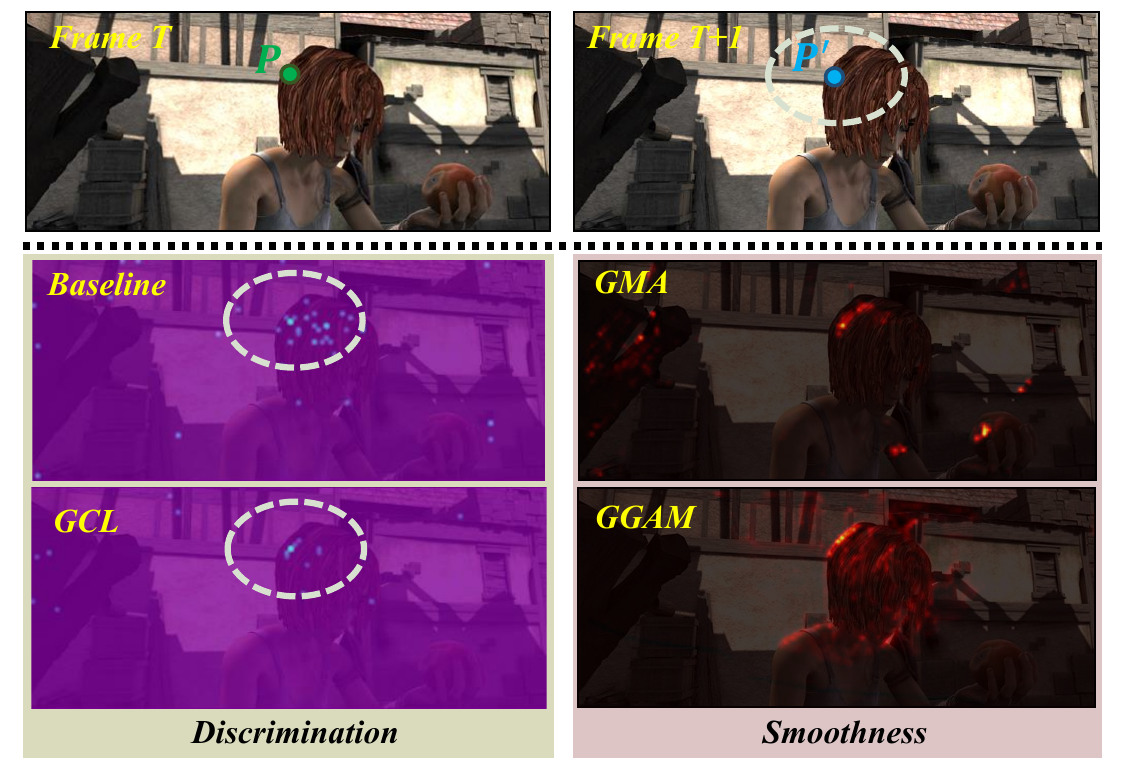}
	\end{center}
	\caption{{\bf Visualization} of feature discrimination (left) and smoothness constraint (right). For a random point $P$ in frame $t$, our GCL enhances feature discrimination by emphasizing fine-grained structural details. Concurrently, our GGAM effectively captures smoothness constraints, centering on pertinent local regions.}
	\label{fig:one}
\end{figure}

In response to the aforementioned challenge, we present the Gaussian Attention Flow network (GAFlow), a pioneering framework that leverages Gaussian Attention (GA) to inform both the feature encoder and the matching module. Central to our approach is a novel attention mechanism equipped with a learnable Gaussian kernel. This mechanism naturally prioritizes the local surroundings of each point, thereby accentuating the importance of local structural information. For the representation learning component, we introduce a novel Gaussian-Constrained Layer (GCL) embedded within the canonical feature encoder. This layer, fortified with a learnable Gaussian kernel, naturally prioritizes the local surroundings of each point, thus accentuating the significance of local structural data. It's worth noting that while the GCL builds upon the pixel-wise modeling typical of the Vision Transformer block, it possesses a sharper focus on local connections, which effectively enhances the feature discrimination for cross-frame matching, as depicted in Fig.~\ref{fig:one}. Moreover, our GCL is dynamic, comprising learnable parameters that can be fine-tuned in concert with the encoder.

In optical flow prediction, traditional techniques prioritize the matching function, viewing it as indispensable when paired with distinctive features. These methods incorporate feature-similarity measures and smoothness constraints to optimize results~\cite{horn1981determining,Brox2004HighAO,Bruhn2005LucasKanadeMH}. With the advent of deep learning, the spotlight has largely shifted to feature-similarity~\cite{huang2022flowformer,Teed2020RAFTRA}, often sidelining the crucial smoothness constraint. When considering it, the implicit smoothness loss is primarily employed~\cite{yu2016back,liu2020arflow}, presuming uniform flow fields —-- a simplification that overlooks intricate object deformations. While some approaches employ Graph techniques~\cite{luo2022learning} or Transformers~\cite{Jiang_2021_ICCV} to capture global motion, they tend to neglect the inherent locality of motion, potentially introducing inaccuracies. The challenge of leveraging neural networks to model motion relationships remains somewhat uncharted.

An ideal neural module for motion modeling should have three key properties: {\bfseries i) Neighbourhood priority}. The motion of object(s) appears locally in visual scenes, and thus the module should instinctively focus more on the nearest neighbor fields for each pixel. {\bfseries ii) Matching-prior awareness}. Previous work~\cite{brox2009large} shows that matching prior helps in large displacements and avoids over-smoothing; {\bf iii) High-order relation centered}. Mining high-order relations~\cite{luo2022learning,Jiang_2021_ICCV,Luo_2022_CVPR} is essential for dealing with occlusions and lighting changes, as low-level similarities (like color) are often fragile. It would also be advantageous if this module's parameters could be trained in a data-driven manner.

Targeting the above goals, we introduce a novel Gaussian-Guided Attention Module (GGAM) to explicitly model the motion affinities for optical flow. To meet the neighborhood priority requirement, our GGAM is formulated as Gaussian-guided attention that naturally emphasizes the neighboring fields of each point. Second, unlike conventional Non-Local operation~\cite{wang2018non}, our GGAM is built across the context feature map and embedded correlation (cost) volume for more comprehensive relation modeling. Specifically, for capturing the matching-prior knowledge, the 4D correlation volumes~\cite{Teed2020RAFTRA} are mapped to be the amplitudes and offsets. The amplitude for each position is set to the Gaussian attention for scaling its amplitude value and the offsets are encoded by the Gaussian attention via the warping operation. It enables free-form deformation of the Gaussian attention and adaptive attention for handling the large displacement of objects. For the last requirement, we draw inspiration from self-attention operations, and map the context feature to the query and key features for modeling the appearance self-similarities. All these operations are fully differentiable in our GGAM. 

Our fully-equipped model, called Gaussian Attention Flow network (GAFlow), unites GCL and GGAM to conduct motion analysis by considering both feature similarities and motion affinities. It achieves the top performance on both Sintel and KITTI benchmarks with limited extra computational cost. Overall, the main {\bf contributions} of this paper are: {\bf 1)} We introduce a novel approach to enhance the local properties of underlying representations. Our proposed Gaussian-Constrained Layer (GCL) can work complementarily with the standard feature encoder to build more discriminative features for optical flow. {\bf 2)} We analyze three important properties for motion affinity modeling, leading to a novel Gaussian-Guided Attention Module (GGAM). For the first time, we show that it is feasible to capture the local relations by learning the Gaussian attention and refining the motion fields for a more reliable optical flow estimation. {\bf 3)} We unite our GCL and GGAM into the contemporary optical flow architecture, making the model stronger at highlighting local structural information. Our Gaussian Attention Flow networks set new records on a variety of benchmarks, \emph{e.g.,} Sintel (clean and final) and KITTI datasets, and outperform existing optical flow models by a relatively large margin.

\begin{figure*}[pt]
	\begin{center}
		\includegraphics[width=0.98\linewidth]{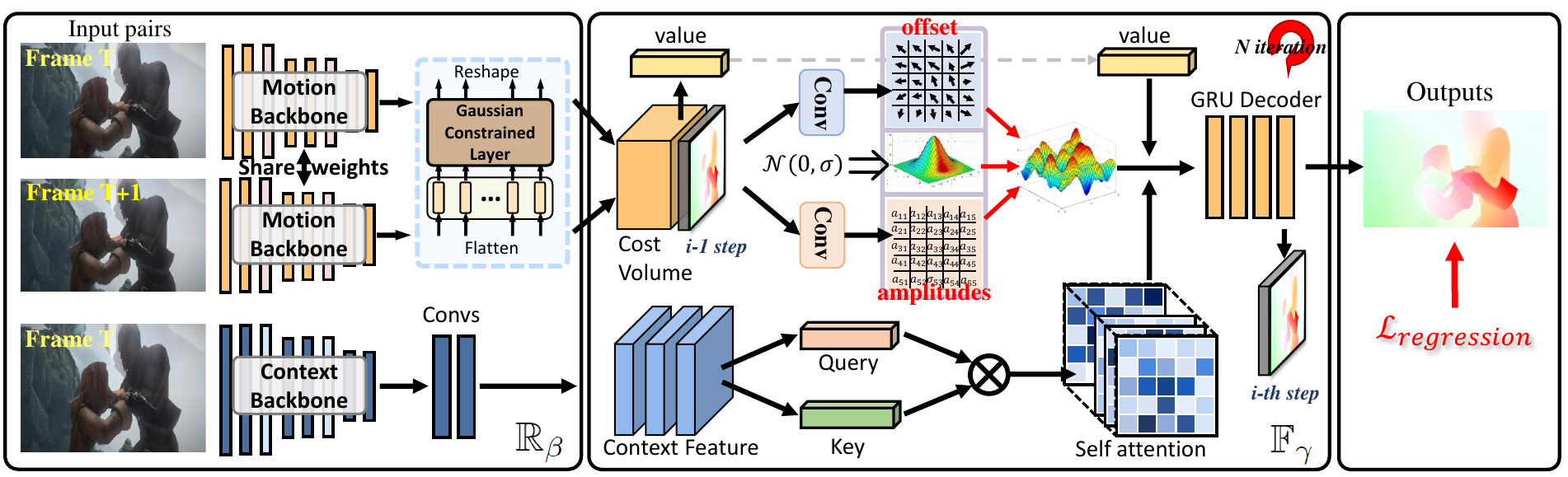}  
	\end{center}
	\caption{{\bf An overview} of the proposed Gaussian Attention Flow network (GAFlow), architected with recurrent learning at its core. During the decoding process, the residual flows are iteratively refined and accumulated to derive the final flow field. ``$\times$'' denotes multiplication. Best viewed in color.}
	\label{fig:2}
\end{figure*}

\begin{figure}[t]
	\begin{center}
		\includegraphics[width=0.99\linewidth]{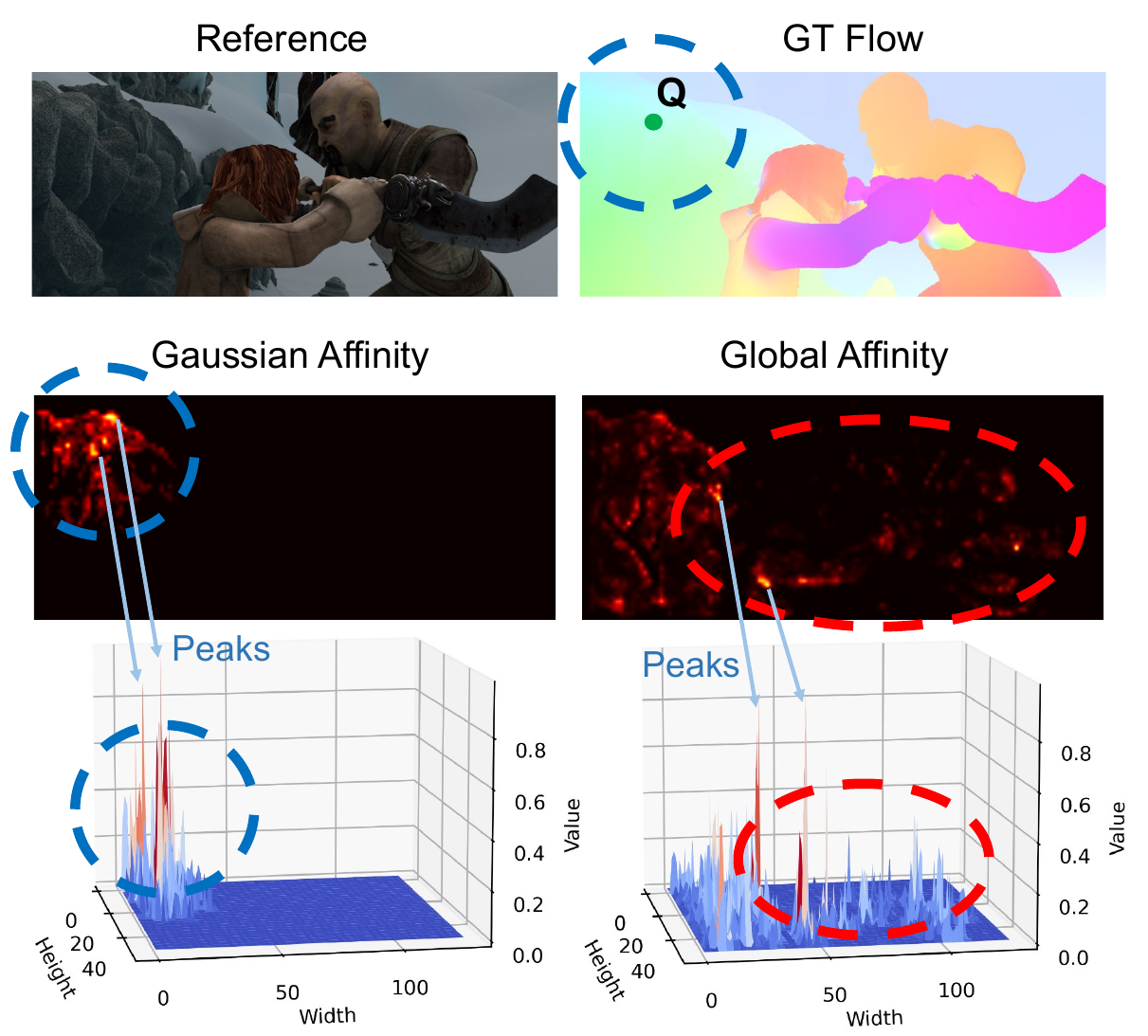}
	\end{center}
	\caption{{\bf A simple illustration} of the Gaussian constraint. ``Q'' represents the query point. The blue circle represents contextual affinity, mirroring the analogous patterns observed in the ground truth flow. Through the Gaussian constraint, we can effectively filter out deceptive relations, as highlighted within the red circle.} 
	\label{fig:motiv}
\end{figure}

\section{Related Work}
\noindent{\bfseries Optical Flow.}
In the epoch of deep learning, early optical flow models, such as \cite{bai2016exploiting,weinzaepfel2013deepflow}, viewed flow field prediction as a mapping challenge, translating two input frames into corresponding flow fields, predominantly harnessing the potency of data. Subsequent models significantly elevated flow estimation accuracy either by adopting more robust representation learning paradigms~\cite{Sun2018PWCNetCF,Hur2019IterativeRR,Zhao2020MaskFlownetAF} or by implementing explicit feature-similarity measurement techniques~\cite{Yang2019VolumetricCN,Wang2020DisplacementInvariantMC}. With the integration of advanced feature learning and cost-volume filtering modules, unsupervised methods have also witnessed remarkable advancements~\cite{luo2021upflow,liu2021asflow,li2023gyroflow}. A notable development is RAFT~\cite{Teed2020RAFTRA}, which employs a 4D correlation volume and a recurrent strategy for optical flow estimation. Building on its recurrent feature-matching paradigm, a slew of contemporary models~\cite{bai2022deep,han2022realflow,luo2023learning,jeong2022imposing,deng2023explicit} have substantially bolstered the dependability of optical flow predictions. To navigate challenges like occlusions and blur effects, innovative tactics—including joint representation learning~\cite{zhao2020msrn}, feature-driven flow regularization~\cite{hui2020lightweight}, iterative refinement~\cite{Hur2019IterativeRR,hui2018liteflownet}, and comprehensive motion analysis~\cite{Jiang_2021_ICCV,luo2022learning}—have been employed. Addressing the smoothness constraint, \cite{Jiang_2021_ICCV} proposed a global motion refinement strategy anchored on attention mechanisms. Meanwhile, \cite{luo2022learning} captures motion affinities using graph techniques and \cite{Luo_2022_CVPR} introduced kernel patch attention (KPA) for the same purpose, though KPA's static kernel window size could potentially curb its efficacy in managing variations in scale and shape.

\noindent{\bfseries Self-Attention.}
Self-attention operations have become instrumental in computer vision for discerning global contexts in images and videos~\cite{wang2018non,wang2020non,zhu2019asymmetric}. Notably, non-local attention, exemplified by \cite{zhu2019asymmetric} and \cite{huang2019ccnet}, is pivotal in visual Transformers~\cite{dosovitskiy2020image,han2022survey}. Recent advancements, such as Swin~\cite{liu2021swin} and NAT~\cite{hassani2023neighborhood}, harness the {\em inductive bias} to enhance Transformers by applying attention within local windows. Inspired by these, our paper presents two Gaussian-based attention modules for optical flow.

\section{Our Approach}
\subsection{Preliminaries}
\noindent {\bf Motivation.}
Fig.~\ref{fig:motiv} visually elucidates the core motivation driving our research. It can be observed that the ground truth optical flow field reveals the intricate nature of motion affinities. These affinities are not only deeply intertwined with contextual relationships but also exhibit pronounced local characteristics. Conventional methods often fall short in aligning with the inherent dynamics of the optical flow field. This observation has propelled us to integrate Gaussian attention into the optical flow model, aiming for a marked enhancement in its precision.

\begin{figure*}
    \centering
    \begin{subfigure}[b]{0.26\textwidth}
        \centering
        \includegraphics[width=\textwidth]{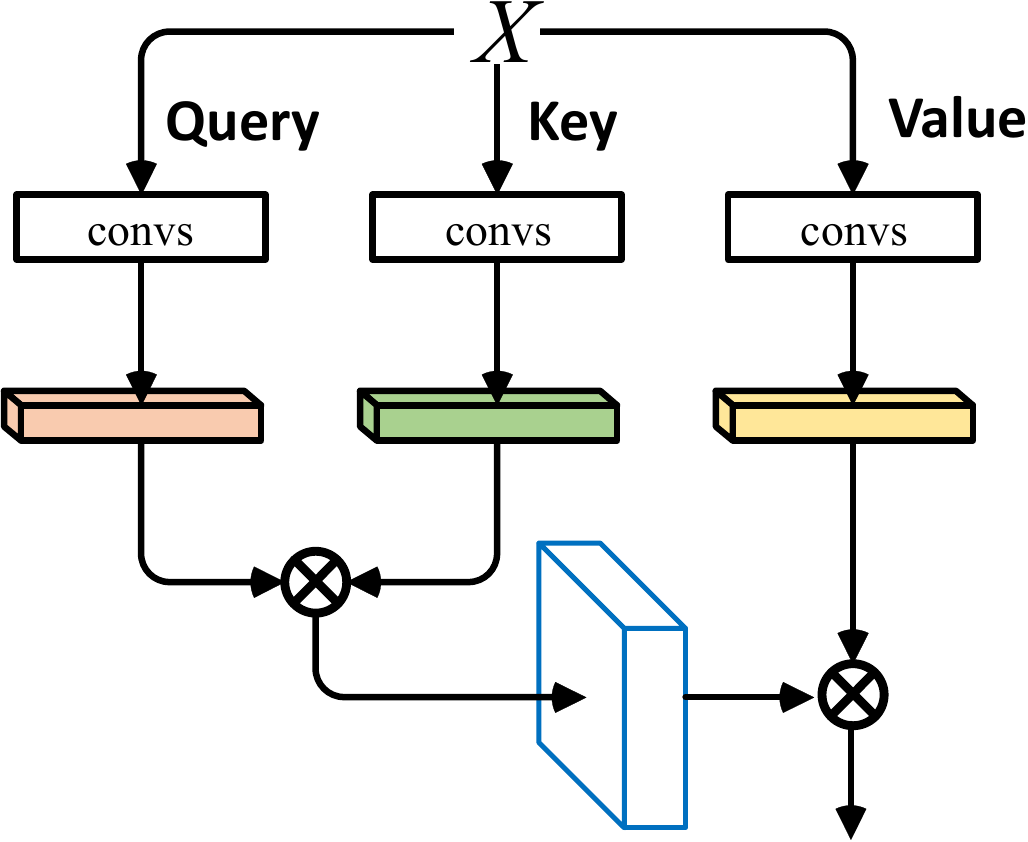}
        \caption{Non-local Attention}
    \end{subfigure}
    \hfill
    \begin{subfigure}[b]{0.30\textwidth}
        \centering
        \includegraphics[width=\textwidth]{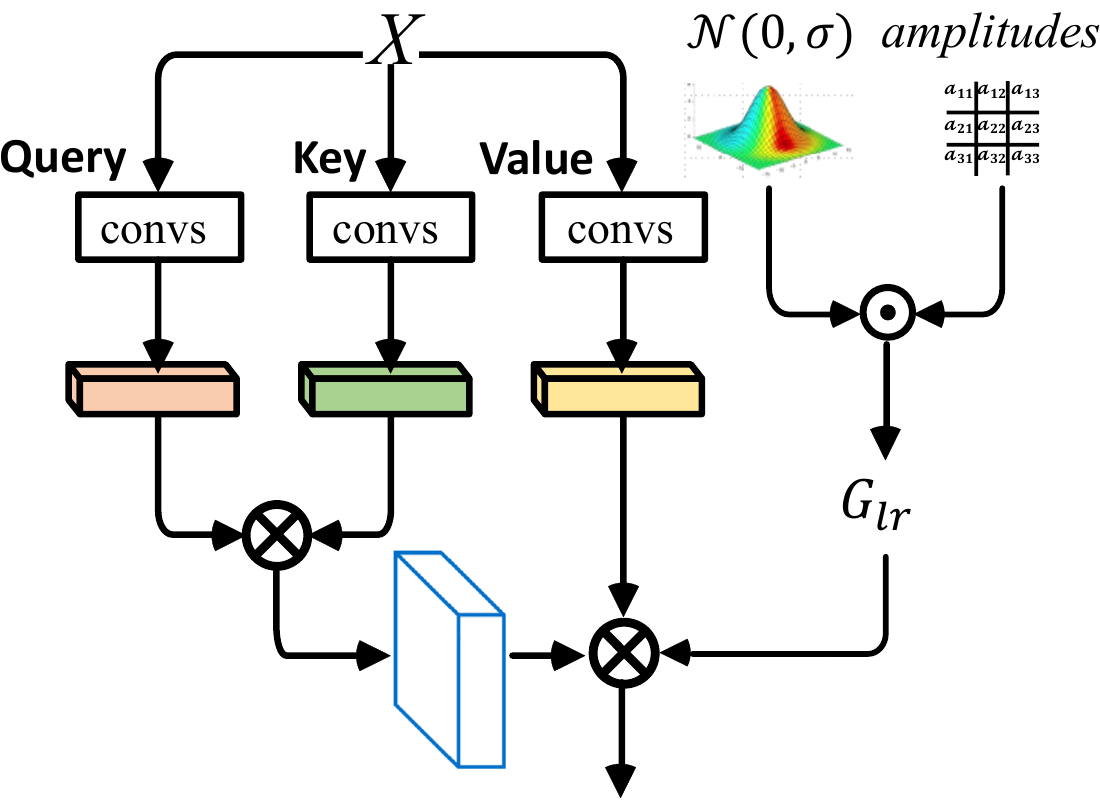}
        \caption{Gaussian Constrained Attention}
    \end{subfigure}
    \hfill
    \begin{subfigure}[b]{0.34\textwidth}
        \centering
        \includegraphics[width=\textwidth]{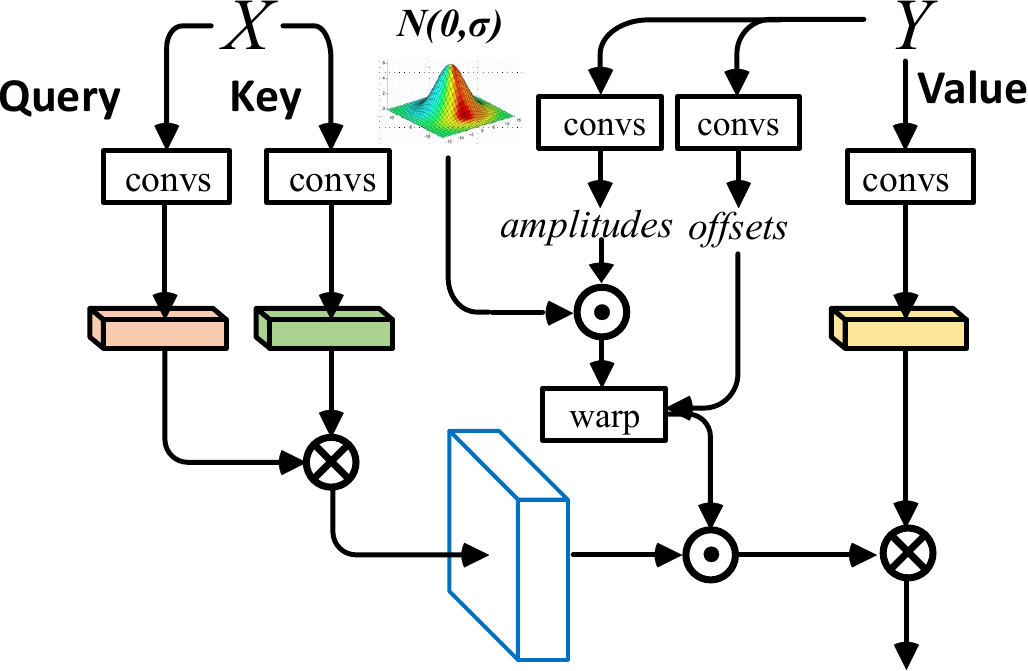}
        \caption{Gaussian Guided Attention}
    \end{subfigure}

	\caption{{\bf Architecture comparisons.} (a) Standard Non-local Attention, (b) Our Gaussian-Constrained Attention tailored for representation enhancement, and (c) Our Gaussian-Guided Attention, designed to model motion affinities using learnable Gaussian attention. Here, {\bf X} represents the basic/context feature, while {\bf Y} signifies the motion feature.}
	\label{fig:1} 
\end{figure*}

\subsection{Overview}
Our GAFlow encompasses two pivotal sub-modules: ~\emph{representation learning} module ($\mathbb{R}_{\beta}$) and \emph{feature matching} module ($\mathbb{F}_{\gamma}$), as visualized in Fig.~\ref{fig:2}. 
Specifically, $\mathbb{R}_{\beta}$ represents the functions of acquiring the feature maps (${\mathbf f}_1, {\mathbf f}_2, {\mathbf f}_c$), corresponding to the feature encoders combined with the proposed Gaussian-Constrained Layer (GCL). $\mathbb{F}_{\gamma}$ pertains to the recurrent flow decoder, fortified with our bespoke Gaussian-Guided Attention Module (GGAM).

At their core, these Gaussian-centric modules can be seamlessly integrated into flow models. Their inclusion accentuates local attributes during both the phases of representation learning and feature matching. Subsequent sections further unpack the mechanics of our GCL and GGAM.

\subsection{Gaussian-Constrained Layer}
The pixel-wise matching is susceptible to feature quality; Yet, the appearance changes caused by the motion blur and the illumination variations lower the feature discrimination. To avoid ambiguities, we propose Gaussian-Constrained Layer (GCL) to obtain the fine-grained structural information and locally discriminative representations for feature matching. Our GCL is designed based upon the standard Transformer~\cite{dosovitskiy2020image, liu2021swin} block (see Fig.~\ref{fig:1} (b)). Specifically, given the base feature $\mathbf x$, it is formulated as:
\begin{equation}
\begin{split}
	&\hat{\mathbf x} = \mathrm{GCA}(\mathrm{LN({\mathbf x})}) + {\mathbf x}, \\
	&{\mathbf y} = \mathrm{FFN}(\mathrm{LN(\hat{\mathbf x})}) + \hat{\mathbf x},
\end{split}
\end{equation}
where $\mathrm{LN}(\cdot)$ and $\mathrm{FFN}(\cdot)$ denote the general layer normalization and feed-forward network respectively in Transformer block. $\mathrm{GCA}(\cdot)$ indicates Gaussian Constrained Attention, which is the core component in our GCL.

Here, we formulate $\mathrm{GCA}$ as a task-specific local operation, which not only avoids global misleading context but also reduces the complexity of attention computation. Therefore, it can be formulated as:
\begin{equation}
\begin{split}
    &{\mathrm Q}_i={L}^Q_i(\mathbf x'), {\mathrm K}_i={L}^K_i(\mathbf x'), {\mathrm V}_i={L}^V_i(\mathbf x'), \\
    &h_i = {\cal G}_{\sf A}({\mathrm Q}_i, \bar {\mathrm K}_i, \bar {\mathrm V}_i),\\
    &{\mathrm H} = \mathrm{Concat}(h_1, h_2, ..., h_n),
\end{split}
\end{equation}
\noindent where ${\mathbf x'} = \mathrm{LN}(\mathbf x)$. ${L}^Q_i(\cdot)$, ${L}^K_i(\cdot)$, and ${L}^V_i(\cdot)$ denote linear projections for the $i$-th head. ${\cal G}_{\sf A}(\cdot)$ is Gaussian attention function, which takes the query feature ${\mathrm Q}_i$, and the regional features of key $\bar {\mathrm K}_i$ and value $\bar {\mathrm V}_i$ as inputs:
\begin{equation}
    {\cal G}_{\sf A}({\mathrm Q}, \bar {\mathrm K}, \bar {\mathrm V}) = \mathrm{Softmax}({\mathrm G}_{lr} \cdot ({\mathrm Q}\bar {\mathrm K}^{\sf T}) / \sqrt{d}) \cdot \bar {\mathrm V},
\end{equation}
where ${\mathrm G}_{lr}$ means a learnable Gaussian kernel with the shape of $k \times k$. It is initialized as a standard Gaussian distribution and can be updated by adding a learnable amplitude matrix $\cal A$ during model training. In the inference process, ${\mathrm G}_{lr}$ works as a constraint mask to reorganize the weights of attentive feature aggregation.

Let the neighborhood of a pixel at the point $p$ be ${\cal N}(p)$; thus the attention on a single pixel can be defined as:
\begin{equation}
    h_{(p)} = \mathrm{Softmax}({\mathrm G}_{lr} \cdot ({\mathrm Q}_{(p)} \bar {\mathrm K})_{{\cal N}(p)}^{\sf T} / \sqrt{d}) \cdot \bar {\mathrm V}_{{\cal N}(p)}.
\end{equation}
Note that the operating range is scalable with the varying region of ${\cal N}(p)$. For instance, it can be extended to all pixels ({\em i.e.,} ${\cal N}(p)$ is equivalent to the image size), leading to a global self-attention in a Gaussian-constrained manner.

\begin{figure*}[t]
	\begin{center}
		\includegraphics[width=0.99\linewidth]{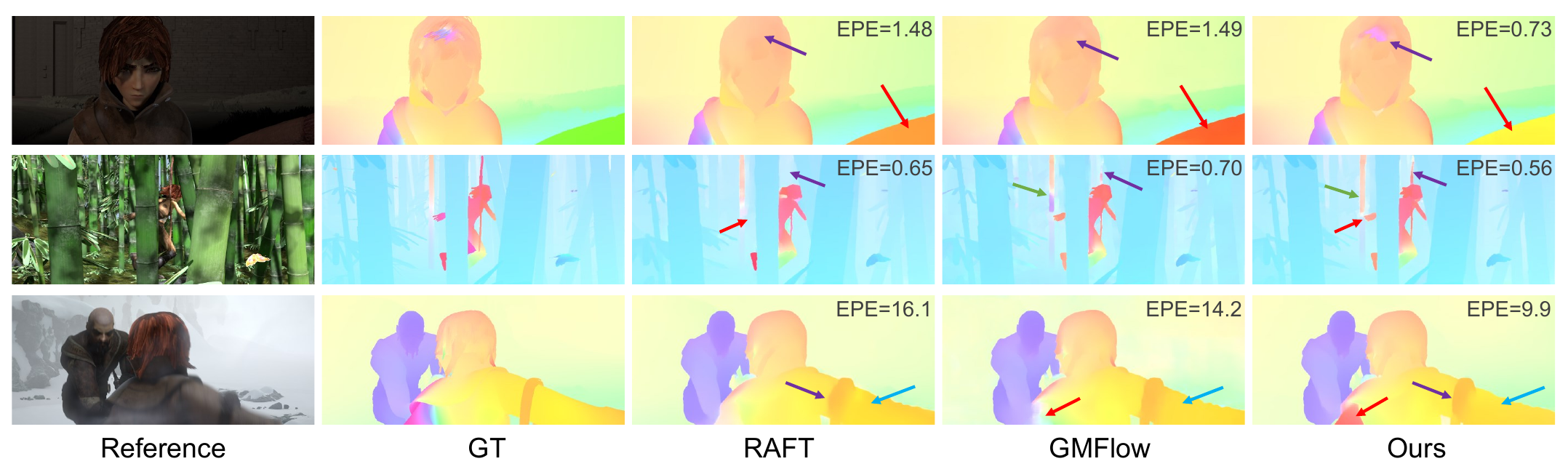}
	\end{center}
	\vspace{-1em}
	\caption{{\bf Qualitative comparisons} with RAFT~\cite{Teed2020RAFTRA} and GMFlow~\cite{xu2022gmflow} on some challenging samples, including large motion, occlusion and motion blur, of Sintel {\tt test} set. The quantitative results are provided by the official website.} 
	\label{fig:sintel}
\end{figure*}

\subsection{Gaussian-Guided Attention Module}
\noindent {\bf Our Idea/Formulation.} In image processing, the Gaussian filter is a widely-used operator that provides a linear scale space for smoothing, where the input image is smoothed at a constant rate in all directions~\cite{witkin1984scale}. Here, our idea is to apply the spatially variant Gaussian to smooth the motion feature ${\mathbf f_m}$. Formally, it can be given as:
\begin{equation}
	({\mathrm F} * {\cal G})(p) = \sum_{p_{ i} \in {\cal N}_p^{}}{{\cal G}(p_{ i} - p) {\mathbf f_m}^{i}},
	\label{eq:gauss}
\end{equation}
where ${\cal N}_p^{}$ denotes a square neighborhood, centered at the pixel $p$, with the pre-defined kernel size $k \times k$, and $p_{i}$ indicates the position ${i}$ in 2D grid space within ${\cal N}_p^{}$. ${\mathbf f_m}^{i}$ refers to the feature at position $i$ within ${\mathbf f_m}$, utilized in the calculation. ${\cal G}{(\cdot)}$ is a 2D Gaussian function, which is defined as:
\begin{equation}
	{\cal G}{(x, y)} = {A}\,\mathrm{exp}(-(\frac{(x-x_0)^2}{2 \sigma_x^2}+\frac{(y-y_0)^2}{2 \sigma_y^2})),
\end{equation}
where ${A}$ denotes the static amplitude (${A} = 1$ as the default setting) and $p_0 = (x_0, y_0)$ means the center point. $(\sigma_x; \sigma_y)$ indicate the variances which are set to $\sigma_x = \sigma_y = \sigma$, determining the operating range of smoothness. However, directly applying the Gaussian smoothing is sub-optimal for capturing the local properties, as the kernel is rigid (or fixed) and the clutters could be inevitably involved in providing false guidance. Next, we show how to make the Gaussian attention more adaptive and also end-to-end trainable.  

\noindent {\bf Gaussian-Guided Attention with Context (GGAC).}
To begin with, we incorporate the contextual information, which preserves the fine-grained structural information, into the Gaussian Attention. Specifically, given the context feature ${\mathbf f}_c \in \mathbb{R}^{c \times h \times w}$ and motion feature ${\mathbf f}_m \in \mathbb{R}^{c \times h \times w}$, we rewrite Eq.~(\ref{eq:gauss}) as:
\begin{equation}
	({\mathrm F} * {\cal G_{\tt C}})(p) = \sum_{p_{ i} \in {\cal N}_p^{}}{{\cal G_{\tt C}}_{(p_{i} - p)}({{\mathbf f}_c}) \rho({{\mathbf f}_m}})^i,
	\label{eq:def}
\end{equation}
where ${\cal G_{\tt C}}_{(p_i - p)}(\cdot)$ is an adaptive kernel function taking the context features in ${\cal N}_p^{}$ as inputs to obtain the context-preserving kernel. $\rho(\cdot)$ is a linear operation for producing the embedded motion feature. We formulate the context-preserving kernel function as:
\begin{equation}
	{\cal G_{\tt C}}_{(p_{ i} - p)}({\mathbf f}_c) = {\cal G'}(p_i - p){\cal F}({\mathbf f}_c),
	\label{eq:ker}
\end{equation}
where ${\cal G'}(p_i - p) \in \mathbb{R}^{k \times k \times h \times w}$ denotes the expanded Gaussian kernel in the spatial dimension, which is prepared for generating the data-dependent kernel matrics with the shape of $k \times k$ based on the corresponding weights on each position in $h \times w$, and ${\cal F}(\cdot)$ is the dynamic weight function for mining the guidance prior on the context feature ${\mathbf f}_c$. 

Rather than simply utilizing a learnable matrix as in convolution kernels, we follow~\cite{Jiang_2021_ICCV} to apply the embedded Gaussian with normalization~\cite{wang2018non} on the context feature $\mathbf f_c$ to produce the attention map. To match the dimension of the pre-defined Gaussian kernel ${\cal G'}(p_i - p) \in \mathbb{R}^{K \times N}$, where $N = h \times w$, and $K = k \times k$ denoting the spatial dimension of kernel size, a naive solution is to discard the redundant information in ${\cal F}_{\tt}({\mathbf f}_c)$. Here, we present a kernel-based context-preserving attention operation, which takes the asymmetric query and key feature maps in the spatial dimension as inputs. In detail, given the embedded context features $\theta({\mathbf f}_c) \in \mathbb{R}^{c \times h \times w}$, we employ an unfolding operator (\eg, torch.nn.Unfold) to get ${\mathrm U}(\theta({\mathbf f}_c))$, thus the dimension is transformed from $c \times N$ to $c \times N \times K$. Then we measure the similarities between all pairs within the kernel region as:
\begin{equation}
	{\cal F}({\mathbf f}_c) = \frac{\mathrm{exp}({{\mathrm U}(\theta({\mathbf f}_c))_l}^T\phi({{\mathbf f}_c})_j / \sqrt{c})}{\sum_{\forall j}{\mathrm{exp}({{\mathrm U}(\theta({\mathbf f}_c))_l}^T\phi({{\mathbf f}_c})_j / \sqrt{c})}},
	\label{eq:cga_attn}
\end{equation}
where $l \in K$ and $j \in c$, and the size of ${\cal F}({\mathbf f}_c)$ is $K \times N$. $\theta(\cdot)$ and $\phi(\cdot)$ indicates linear projections for channel-wise feature aggregation~\cite{Jiang_2021_ICCV}.

\begin{table*}[t]
	\centering
	\begin{spacing}{1.05}
	\resizebox{0.99\textwidth}{!}{
		\begin{tabular}{cllccccccc}
			\toprule
			\multirow{2}{2.2cm}{\centering Training} & \multirow{2}{3.0cm}{\centering{Method}} & \multirow{2}{2.2cm}{Reference} & \multicolumn{2}{c}{\underline{Sintel (\tt Val)}} &  \multicolumn{2}{c}{\underline{KITTI-15 (\tt Val)}} & \multicolumn{2}{c}{\underline{Sintel (\tt Test)}} & \multicolumn{1}{c}{\underline{KITTI-15 (\tt Test)}} \\
			& & & Clean & Final & EPE & F1-all & Clean & Final & F1-all \\
			\midrule    
			\multirow{14}{2.5cm}{~~~{\tt Val}: C + T / \qquad {\tt Test}: +S+K+H} 
			& RAFT~\cite{Teed2020RAFTRA} & ECCV-20      & 1.43 & {2.71} & {5.04} & {17.4} & {1.61} & {2.86} & {5.10} \\
			& SCV~\cite{Jiang2021LearningOF} & CVPR-21 & {1.29} & 2.95 & 6.80 & 19.3 & 1.77 & 3.88 & 6.17 \\
			& GMA~\cite{Jiang_2021_ICCV} & ICCV-21 & {1.30} & 2.74 & 4.69 & 17.1 & 1.39 & 2.47 & 5.15 \\
			& SeparableFlow\cite{zhang2021separable} & ICCV-21 & {1.30} & 2.59 & 4.60 & 15.9 & 1.50 & 2.67 & 4.64 \\
			& Flow1D~\cite{xu2021high} & ICCV-21 & 1.98 & 3.27 & 6.69 & 23.0 & 2.24 & 3.81 & 6.27 \\
			& AGFlow~\cite{luo2022learning} & AAAI-22 & 1.31 & 2.69 & 4.82 & 17.0 & 1.43 & 2.47 & 4.89 \\
			& GMFlowNet~\cite{zhao2022global} & CVPR-22 & 1.14 & 2.71 & \underline {4.24} & 15.4 & 1.39 & 2.65 & 4.79 \\
			& GMFlow~\cite{xu2022gmflow} & CVPR-22 & \underline {1.08} & 2.48 & 7.77 & 23.4 & 1.74 & 2.90 & 9.32 \\
			& CRAFT~\cite{sui2022craft} & CVPR-22 & 1.27 & 2.79 & 4.88 & 17.5 & 1.45 & 2.42 & 4.79 \\
			& DIP~\cite{zheng2022dip} & CVPR-22 & 1.30 & 2.82 & 4.29 & \bf 13.7 & 1.44 & 2.83 & \bf 4.21 \\
			& KPA-Flow~\cite{luo2022learning} & CVPR-22 & 1.28 & 2.68 & 4.46 & 15.9 & 1.35 & 2.36 & 4.60 \\
			& OCTC~\cite{jeong2022imposing} & CVPR-22 & 1.31 & 2.67 & 4.72 & 16.3 & 1.41 & 2.57 & \underline {4.33} \\
			& SKFlow~\cite{sun2022skflow} & NeurIPS-22 & 1.22 & \underline {2.46} & 4.27 & 15.5 & \underline {1.28} & \underline {2.27} & 4.84 \\
			& \bf {GAFlow (ours)} & & \bf 1.02 & \bf 2.45 & \bf 3.98 & \underline {15.0} & \bf 1.21 & \bf 2.24 & 4.52 \\

			\midrule
			\multirow{2}{2.0cm}{A / +TSKHV}
			& RAFT-A~\cite{sun2021autoflow} & CVPR-21  & 1.95 & 2.57 & 4.23 & - & 2.01 & 3.14 & 4.78 \\
			& RAFT-it~\cite{sun2022disentangling} & ECCV-22  & 1.74 & 2.41 & 4.18 & \bf 13.4 & 1.55 & 2.90 & \bf 4.31 \\
			\multirow{2}{2.0cm}{I+CT / +SKH} & FlowFormer~\cite{huang2022flowformer} & ECCV-22 & \underline{1.01} & \underline{2.40} & \underline{4.09} & 14.7 & \underline{1.16} & \underline{2.09} & 4.68 \\
			& \bf GAFlow-FF (ours) & &  \bf 0.95 & \bf 2.34 & \bf{3.92} & \underline{13.9} & \bf 1.15 & \bf 2.05 & \underline{4.42} \\
			\bottomrule
		\end{tabular}
	}
	\end{spacing}
	\caption{{\bf Quantitative comparison} with state-of-the-art models on standard benchmarks for cross-dataset evaluation and online testing. ``C+T'' indicates the models trained on FlyingChairs and FlyingThings for generalization ability evaluation. ``+S+K+H'' denotes more training data involved from Sintel, KITTI-2015, and HD1K. ``A'' indicates the models are trained on AutoFlow~\cite{sun2021autoflow} dataset for {\tt Val} set evaluation. ``V'' denotes VIPER~\cite{richter2017playing} dataset. ``I'' means the encoders are pre-trained on ImageNet~\cite{deng2009imagenet}.
    The best and second-best results are marked in {\bf bold} and \underline{underline}, respectively.
	} \label{tab:1}
\end{table*}

\noindent {\bf Gaussian-Guided Attention with Deformation (GGAD).}
While Gaussian attention with context effectively addresses the problem of flow field over-smoothing caused by the data-agnostic Gaussian kernel, it still faces a limitation due to the rigid operating region of the Gaussian kernel. This constraint hinders the smoothness performance for motion feature refinement. 
To mitigate this issue, we take a further step to optimize the operating region of the Gaussian kernel by learning the deformable kernels in a data-driven manner, as shown in Fig.~\ref{fig:1} (c). To avoid the static amplitude $A$ of the Gaussian function, we design a dynamic amplitude operator based on the cross-frame matching features.
In practice, similar to Eq.~(\ref{eq:def}), our Gaussian-guided attention can be further formulated as:
\begin{equation}
	({\mathrm F} * {\cal G_{\tt D}})(p) = \sum_{p_{ i} \in {\cal N}_p^{}}{{\cal G_{\tt D}}_{(p_{ i} - p)}({\mathbf f}_c, {\mathbf f_m}) \rho({{\mathbf f}_m}}),
	\label{eq:dga}
\end{equation}
where ${\cal G_{\tt D}}_{(p_{ i} - p)}(\cdot)$ is the deformable Gaussian kernel function that is used to produce the more flexible Gaussian attention for adaptively capturing the relevant information and avoiding rigid kernel boundary in a data-dependent manner. Specifically, the deformable Gaussian kernel is given by:
\begin{equation}
	{\cal G_{\tt D}}_{(p_{ i} - p)}({\mathbf f}_c, {\mathbf f_m}) = {\cal A}({\mathbf f}_{m}){\cal D}_{(p_i - p)}({\mathbf f_m}){\cal F}({\mathbf f}_c),
	\label{eq:ker_dga}
\end{equation}
where ${\cal A}(\cdot)$ indicates the amplitude operator taking the motion feature $\bf{f}_m$ to produce adaptive amplitudes, ${\cal F}({\mathbf f}_c)$ is the kernel-based context-preserving attention matrix as in Eq.~(\ref{eq:cga_attn}), and ${\cal D}_{(p_i - p)}(\cdot)$ denotes the deformable Gaussian operator, which can be formulated as:
\begin{equation}
	{\cal D}_{(p_i - p)}({\mathbf f_m}) = {\cal W}({\cal G'}(p_i - p), {\cal O}({\mathbf f_m})),
\end{equation}
where ${\cal G'}(p_i - p) \in \mathbb{R}^{k \times k \times h \times w}$ indicates the spatially expanded Gaussian kernel as in Eq.~(\ref{eq:ker}), and ${\cal O}({\mathbf f_m})$ denotes the offset maps predicted from the motion feature. ${\cal W}(\cdot)$ is the warp function taking the obtained offsets as inputs and operating on the Gaussian kernel. 

Another important component in the deformable Gaussian kernel (Eq.~(\ref{eq:ker_dga})) is the amplitude operator ${\cal A}(\cdot)$, which is given by:
\begin{equation}
	{\cal A}({\mathbf f}_{m}) = 1 + \vartheta({\mathbf f}_{m})\lambda,
\end{equation}
where $\lambda$ indicates a learnable parameter to constrain the fluctuation of the generated amplitude, and $\vartheta(\cdot)$ is a linear function. Finally, similar to the expanding strategy in the deformable Gaussian operator, the learned amplitude ${\cal A}({\mathbf f}_{m})$ can be easily transformed to operate on the  Gaussian kernel ${\cal G'}(p_i - p)$ for generating a more flexible one.

\section{Experimental Results}

\subsection{Implementation Details}

For fair comparison with existing methods~\cite{Jiang_2021_ICCV, sui2022craft, Luo_2022_CVPR}, we first plug $1$ GCL and $1$ GGAM into RAFT~\cite{Teed2020RAFTRA} to build a small model, {\em i.e.,} GAFlow-S. Specifically, Gaussian-Constrained Layer (GCL) is appended to the motion encoder with $\sigma = 9$. Gaussian-Guided Attention Module (GGAM) performs feature smoothing on context and motion features, where we set sigma to $20$ and $15$ for Sintel and KITTI, respectively. 
Besides, to further explore the potential of our approach, we also employ a stronger baseline with POLA~\cite{zhao2022global} modules and SKBlocks~\cite{sun2022skflow}, which is regarded as the default model of GAFlow.
Furthermore, we enhance the model's capabilities by integrating the proposed module into FlowFormer~\cite{huang2022flowformer}, resulting in a robust and advanced model termed GAFlow-FF.

All experiments are conducted based on PyTorch toolbox. During training our GAFlow model, we follow RAFT to set the batch sizes to $6$ and adopt AdamW optimizer with one-cycle learning rate policy~\cite{Teed2020RAFTRA}. Similar to previous works~\cite{Teed2020RAFTRA,Jiang_2021_ICCV, luo2022learning}, the synthetic datasets are also involved in model pre-training. The pre-training iterations on FlyingChairs~\cite{Dosovitskiy2015FlowNetLO} and FlyingThings~\cite{Mayer2016ALD} are set to 120k and 160k, respectively. Then the fine-tuning for Sintel online evaluation is conducted on the combined training data of Sintel~\cite{Butler2012}, KITTI-2015~\cite{KITTI_2015}, and HD1K~\cite{Kondermann2016TheHB} for 160k iterations. Finally, the trained model requires another 50k iterations of finetuning on data of KITTI-2015~\cite{KITTI_2015} before KITTI testing. We use only a single GPU and set the batch size to $1$ for evaluation and online testing.

\subsection{Comparison with State-of-the-Arts}

\noindent{\bf{Results on Sintel.}}
We first compare the proposed approach with state-of-the-art methods for generalization evaluation on Sintel. As shown in {\tt Val} sets of Tab.~\ref{tab:1}, our approach consistently achieves the best performance on all metrics. Specifically, GAFlow achieves the best EPE scores at $1.02$ and $2.45$ on the ``C + T'' setting. 
Besides, GAFlow-FF improves the SOTA performance to $0.95$ and $2.34$ in EPE on Sintel clean and final passes, respectively. 
For online testing, we follow prior works~\cite{Jiang_2021_ICCV,Luo_2022_CVPR,sui2022craft} to utilize the warm-start strategy~\cite{Teed2020RAFTRA}. GAFlow achieves the best EPE scores at $1.21$ and $2.24$ on the standard training setting.  
Moreover, GAFlow-FF sets a new record at $1.15$ and $2.05$ in EPE, outperforming recent works by a relatively large margin.

We compare the proposed GAFlow with the well-known methods RAFT~\cite{Teed2020RAFTRA} and GMFlow~\cite{xu2022gmflow} on Sintel {\tt test} set, and some qualitative comparisons are illustrated in Fig.~\ref{fig:sintel}. The results demonstrate that the proposed approach can effectively leverage the local structural information and high-order relations to resolve the ambiguities in motion learning, leading to a more robust and flexible model for handling the challenges in optical flow estimation.

\noindent{\bf{Results on KITTI.}}
As shown in Tab.~\ref{tab:1}, GAFlow achieves $3.98$ in EPE and $15.0\%$ in F1-all on the KITTI {\tt Val} set, which is comparable with the SOTA model FlowFormer yet with less computational overhead. Moreover, GAFlow-FF further boosts the scores to $3.92$ and $13.9\%$. For online testing, our models achieve the top-ranked performances at $4.52\%$ and $4.42\%$ in F1-score.

\begin{figure}[t]
	\begin{center}
		\includegraphics[width=0.99\linewidth]{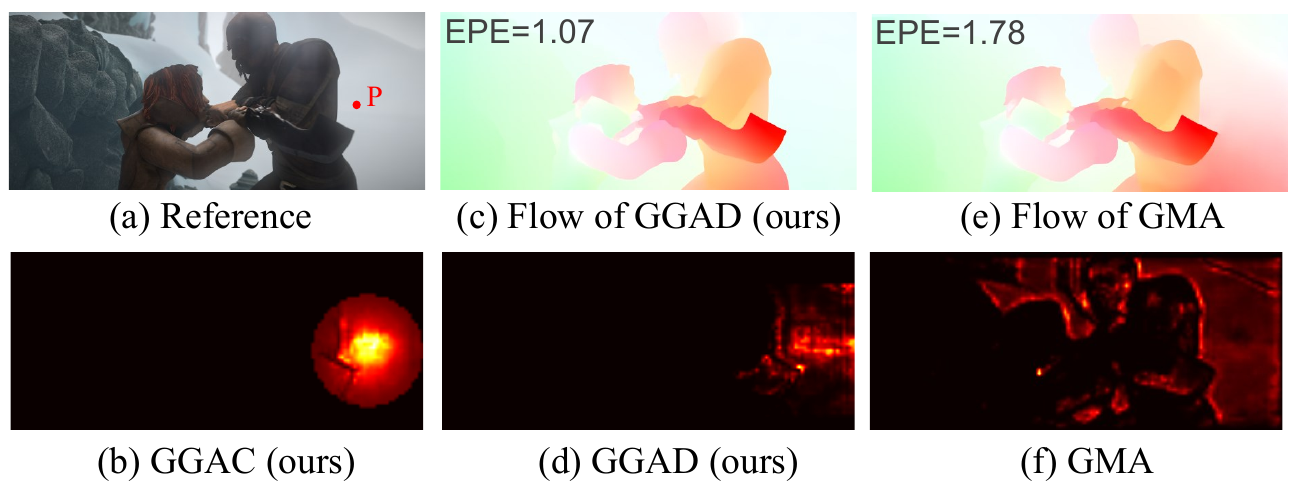}
	\end{center}
	\vspace{-.5em}
	\caption{{\bf A challenging sample} from the Sintel final pass. The second row provides attention maps of our GGAC, GGAD and GMA~\cite{Jiang_2021_ICCV}, respectively. ``P'' indicates the query point to obtain attention maps.}
	\label{fig:attn}
\end{figure}

\subsection{Ablation Analysis} 
\label{sec:4}

\noindent{\bf Comparison with GMA.}
GMA~\cite{Jiang_2021_ICCV} presents a global strategy for handling occlusion in flow estimation. As shown in Tab.~\ref{tab:cmpr_small}, our GAflow-S achieves better performance on all metrics. This is because the global attention on context features involves too much misleading information for motion guidance, which contains long-range similarities in appearance but is improper for motion refinement (see Fig.~\ref{fig:attn} (f)). In contrast, our approach effectively takes the local property of regional context affinities into consideration, and leverages the extracted high-order relations with the flexible operating range to better infer the flow fields.

\noindent{\bf Comparison with KPA.}
The recent work KPA-Flow~\cite{Luo_2022_CVPR} shares a similar insight with our approach, leveraging local attention to avoid global misleading information. However, its intrinsic defects in formulation inevitably hinder the effect of motion refinement. First, the zero padding patches (out of the image) lead to obvious deviation for similarity measurement. Second, the neighboring pixels could be partitioned into distinct pre-defined (fixed) patch and kernel windows, leading to abrupt shifts in operating regions. In contrast, our approach enables a flexible constraint and freeform operating regions in a learnable manner, ensuring the proper and stable guidance pattern. Moreover, we provide quantitative comparisons in Tab.~\ref{tab:cmpr_small}, where our approach achieves better performance on all metrics.

\begin{table}[t]
	\setlength{\tabcolsep}{0.35\tabcolsep}
	\centering
	\resizebox{0.45\textwidth}{!}{
	  \begin{tabular}{ll*{4}{c}}
	  \toprule
	  \multirow{2}{*}{\centering Sintel} & \multirow{2}{*}{\centering Type} & RAFT\cite{Teed2020RAFTRA} & GMA\cite{Jiang_2021_ICCV} & KPA\cite{Luo_2022_CVPR} & \bf GAFlow-S \\
	  & & (EPE) & (EPE) & (EPE) & (EPE) \\
	  \midrule
	  & s0-10 & 0.32 & 0.27 & 0.26 & \bf 0.24 \\
	  Clean & \emph{s}10-40 & 1.55 & 1.35 & 1.38 & \bf 1.29 \\
	  (train) & \emph{s}40+ & 8.83 & 8.55 & 8.43 & \bf 7.92 \\
	  & All & 1.41 & 1.31 & 1.28 & \bf 1.21 \\
	  \midrule
	  & \emph{s}0-10 & 0.52 & 0.54 & 0.54 & \bf 0.51 \\
	  Final& \emph{s}10-40 & 3.00 & 3.07 & 3.02 & \bf 2.83 \\
	  (train)& \emph{s}40+ & 17.45 & 17.57 & 17.34 & \bf 17.02 \\
	  & All & 2.69 & 2.74 & 2.68 & \bf 2.56 \\
	  \bottomrule
	  \end{tabular}
	}
  	\caption{{\bf Quantitative results} on different displacements. All methods are trained on ``C+T'' and evaluated on Sintel for fair comparison. \emph{s}0-10, \emph{s}10-40 and \emph{s}40+ denote the ground truth flow value belonging to 0-10, 10-40 and 40+ pixels~\cite{Butler2012}, respectively.}
	\label{tab:cmpr_small}
\end{table}

\noindent{\bf Ablation for Feature Enhancement.}
In Tab.~\ref{tab:cmpr} (\# 1), we first compare the proposed GCL with the widely recognized Transformer blocks~\cite{liu2021swin,zhao2022global,qin2022cosformer}, which have been extensively demonstrated to be effective for feature extraction.
Swin Transformer~\cite{liu2021swin} requires at least two blocks to perform the window shifting strategy, consuming more computation overhead. For other methods, only one block is applied for fair comparison. 
POLA~\cite{zhao2022global} has been shown to be an effective local attention-based approach for feature extraction with $6$ or $12$ stacked blocks (requiring extra $4.35/8.7$ M parameters). As shown in the table, using one block (POLA$_{\times 1}$) significantly weaken the power. 
The advanced linear Transformer cosFormer~\cite{qin2022cosformer} exhibits a lower EPE score on Sintel Clean, whereas our GCL performs better on other metrics.
Besides, we conduct an extra ablation for using a learnable kernel without Gaussian (L.Kernel). As can be seen, our GCL outperforms L.Kernel across all metrics. Particularly on the challenging Sintel, without a Gaussian constraint, performance declines significantly due to the uncontrolled misleading similarities at the kernel border.
All experiments illustrate that, requiring similar computational overhead, the proposed GCL surpasses other methods by a relatively large margin.

\noindent{\bf Ablation for Stronger Encoder.}
The smoothness constraint for flow estimation can be naturally regarded as a complementary component to the feature matching process, and benefits from stronger encoders for feature extraction. Thus, to further explore the potential of our approach, we follow prior work~\cite{zhao2022global} to employ a stronger motion encoder with the standard POLA blocks, which largely boosts the performance on Sintel clean pass, as in Tab.~\ref{tab:cmpr} (\# 2). 
Inspired by GMFlowNet~\cite{zhao2022global}, we build two types of encoding modules for separately handling Sintel and KITTI datasets.

\begin{table}[t]
	\centering
	\setlength{\tabcolsep}{5pt}
	\resizebox{0.47\textwidth}{!}{
        \begin{tabular}{llcccc}
            \toprule
            \multirow{2}{*}{\centering Method} &  \multicolumn{2}{c}{\centering Sintel (\tt Val)} & \multicolumn{2}{c}{\centering KITTI-15 (\tt Val)} & \multirow{2}{0.8cm}{\centering Param.} \\
            \cmidrule(lr){2-3}
         	\cmidrule(lr){4-5}
            & Clean & Final & EPE & F1-all & \\
			\midrule
			RAFT~\cite{Teed2020RAFTRA} & 1.43 & 2.71 & 5.04 & 17.4 & 5.3M  \\

			\midrule
			\multicolumn{6}{c}{\centering \# 1: Feature Enhancement} \\
			Swin Trans.~\cite{liu2021swin} & 1.45 & 2.75 & 5.16 & 17.3 & 7.3M \\
			POLA$_{\times 1}$~\cite{zhao2022global} & 1.42 & 2.68 & 4.91 & 16.9 & 6.5M \\
			cosFormer~\cite{qin2022cosformer} & 1.29 & 2.82 & 4.95 & 17.3 & 6.7M \\
			L.Kernel & 1.42 & 2.76 & 4.69 & 16.6 & 6.5M \\
			\underline{GCL} & 1.33 & 2.67 & 4.61 & 16.5 & 6.5M \\

			\midrule
			\multicolumn{6}{c}{\centering \# 2: Stronger Encoder} \\
			\underline{POLAs~\cite{zhao2022global}} & 1.18 & 2.63 & 4.52 & 16.6 & 10.0M \\
			No & 1.33 & 2.67 & 4.61 & 16.5 & 6.5M \\

			\midrule
			\multicolumn{6}{c}{\centering \# 3: Smoothing Pattern} \\
			+ GGAC & 1.13 & 2.62 & 4.33 & 15.8 & 10.1M \\
			\underline{+ GGAD} & 1.08 & 2.56 & 4.26 & 15.6 & 10.1M \\

			\midrule
			\multicolumn{6}{c}{\centering \# 4: Compatibility with SOTA modules} \\
			Refine.Sp4~\cite{xu2022gmflow} & 1.03 & 2.45 & 4.10 & 14.6 & 12.5M \\ 
			\underline{SKBlocks~\cite{sun2022skflow}} & 1.02 & 2.45 & 3.98 & 15.0 & 10.4M \\
            
            \bottomrule
        \end{tabular}
	}
	\caption{{\bf Ablation experiments.} L.Kernel indicates the learnable kernel without Gaussian. Settings as default are \underline{underlined}. Refer to Sec.~\ref{sec:4} for detailed comparisons.}
	\label{tab:cmpr}
\end{table}

\begin{table}[t]
	\centering
	\resizebox{0.47\textwidth}{!}{
        \begin{tabular}{lcccc}
            \toprule
            \multirow{2}{*}{\centering Method} & \multicolumn{2}{c}{\centering KITTI-15 (\tt Val)} & \multirow{2}{1.2cm}{\centering Param. } & \multirow{2}{1.2cm}{Time (s)} \\
            \cmidrule(lr){2-3}
            & EPE & F1-all &  & \\
			\midrule
			GMFlowNet~\cite{zhao2022global} & 4.24 & 15.4 & 9.3M & 0.32 \\
			FlowFormer~\cite{huang2022flowformer} & 4.09 & 14.7 & 18.2M & 1.77 \\
			\bf GAFlow (ours) & 3.98 & 15.0 & 11.1M & 0.36 \\
			\bf GAFlow-FF (ours) & 3.92 & 13.9 & 18.2M & 1.85 \\

			\bottomrule
		\end{tabular}
	}
	\caption{Quantitative comparisons for computational complexity. The input size is $376 \times 1248$ as in KITTI dataset.}
	\label{tab:time}
\end{table}

\noindent{\bf Ablation for Smoothing Pattern.}
The proposed GGAC is capable of producing flexible weight matrices for motion feature smoothing, which effectively tackles the issue of flow field over-smoothing caused by the data-agnostic Gaussian kernel. However, the issues of rigid operating region and static amplitude still affect the motion refinement (see Fig.~\ref{fig:attn} (b) and Tab.~\ref{tab:cmpr} (\# 3)). Thus we further improve the approach with the spatially deformable kernel with the dynamic amplitude in a data-driven manner, \ie, GGAD, which boosts the performance by around $~3\%$. 

\noindent{\bf Ablation for Compatibility with SOTA modules.}
In Tab.~\ref{tab:cmpr} (\# 4), similar to prior work~\cite{xu2022gmflow}, we further perform flow refinement with the flow decoder on 1/4 features, termed Refine.Sp4, which helps to boost the performance on both Sintel and KITTI datasets. Moreover, we follow SKFlow~\cite{sun2022skflow} to employ SKBlocks in the motion encoding and state updating modules of the flow decoder. The ablation indicates that the GGAD module is compatible with SKBlocks, and the integration of the decoder leads to improved performance across all metrics, while introducing minimal additional parameters ($10.1 \rightarrow 11.1$ M).

\noindent {\bf Runtime Comparison.}
The comparisons of KITTI evaluation results, parameters and runtime are presented in Tab.~\ref{tab:time}. Although FlowFormer~\cite{huang2022flowformer} uses ImageNet for encoder pre-training, our GAFlow achieves a competitive performance (even better in online testing, see Tab.~\ref{tab:1}) while consuming fewer parameters by $7.1$ M and reducing the inference time by $80.0\%$. Moreover, built on the previous SOTA model FlowFormer, our GAFlow-FF largely reduces the error by $4.8\%$ with negligible computational complexity.

\section{Conclusion}
In this work, we introduce a novel Gaussian Attention Flow network (GAFlow) to explicitly highlight local properties during representation learning and enforce the motion affinity during matching. We deliver two Gaussian-based modules, \ie Gaussian-Constrained Layer (GCL) and Gaussian-Guided
Attention Module (GGAM), which works well with existing flow architectures and can greatly enhance the reliability of optical flow estimation. Extensive quantitative and qualitative comparisons performed on commonly-used datasets demonstrate that our method significantly outperforms the current alternatives.

\noindent\textbf{Acknowledgements}. This work is supported by Sichuan Science and Technology Program of China under grant No.2023NSFSC0462.

{\small
\bibliographystyle{ieee_fullname}
\bibliography{egbib}

\begin{thebibliography}{10}\itemsep=-1pt

\bibitem{amiranashvili2018motion}
Artemij Amiranashvili, Alexey Dosovitskiy, Vladlen Koltun, and Thomas Brox.
\newblock Motion perception in reinforcement learning with dynamic objects.
\newblock In {\em CRL}, 2018.

\bibitem{bai2016exploiting}
Min Bai, Wenjie Luo, Kaustav Kundu, and Raquel Urtasun.
\newblock Exploiting semantic information and deep matching for optical flow.
\newblock In {\em ECCV}, 2016.

\bibitem{bai2022deep}
Shaojie Bai, Zhengyang Geng, Yash Savani, and J~Zico Kolter.
\newblock Deep equilibrium optical flow estimation.
\newblock In {\em CVPR}, 2022.

\bibitem{brox2009large}
Thomas Brox, Christoph Bregler, and Jitendra Malik.
\newblock Large displacement optical flow.
\newblock In {\em CVPR}, 2009.

\bibitem{Brox2004HighAO}
T. Brox, Andr{\'e}s Bruhn, N. Papenberg, and J. Weickert.
\newblock High accuracy optical flow estimation based on a theory for warping.
\newblock In {\em ECCV}, 2004.

\bibitem{Bruhn2005LucasKanadeMH}
Andr{\'e}s Bruhn, J. Weickert, and C. Schn{\"o}rr.
\newblock Lucas/kanade meets horn/schunck: combining local and global optic flow methods.
\newblock {\em IJCV}, 2005.

\bibitem{Butler2012}
Daniel Butler, Jonas Wulff, Garrett Stanley, and Michael Black.
\newblock A naturalistic open source movie for optical flow evaluation.
\newblock In {\em ECCV}, 2012.

\bibitem{deng2023explicit}
Changxing Deng, Ao Luo, Haibin Huang, Shaodan Ma, Jiangyu Liu, and Shuaicheng Liu.
\newblock Explicit motion disentangling for efficient optical flow estimation.
\newblock In {\em ICCV}, 2023.

\bibitem{deng2009imagenet}
Jia Deng, Wei Dong, Richard Socher, Li-Jia Li, Kai Li, and Li Fei-Fei.
\newblock Imagenet: A large-scale hierarchical image database.
\newblock In {\em CVPR}, 2009.

\bibitem{dosovitskiy2020image}
Alexey Dosovitskiy, Lucas Beyer, Alexander Kolesnikov, Dirk Weissenborn, Xiaohua Zhai, Thomas Unterthiner, Mostafa Dehghani, Matthias Minderer, Georg Heigold, Sylvain Gelly, et~al.
\newblock An image is worth 16x16 words: Transformers for image recognition at scale.
\newblock In {\em ICLR}, 2021.

\bibitem{Dosovitskiy2015FlowNetLO}
A. Dosovitskiy, P. Fischer, Eddy Ilg, Philip H{\"a}usser, Caner Hazirbas, V. Golkov, P.~V.~D. Smagt, D. Cremers, and T. Brox.
\newblock Flownet: Learning optical flow with convolutional networks.
\newblock In {\em ICCV}, 2015.

\bibitem{han2022survey}
Kai Han, Yunhe Wang, Hanting Chen, Xinghao Chen, Jianyuan Guo, Zhenhua Liu, Yehui Tang, An Xiao, Chunjing Xu, Yixing Xu, et~al.
\newblock A survey on vision transformer.
\newblock {\em TPAMI}, 2022.

\bibitem{han2022realflow}
Yunhui Han, Kunming Luo, Ao Luo, Jiangyu Liu, Haoqiang Fan, Guiming Luo, and Shuaicheng Liu.
\newblock Realflow: Em-based realistic optical flow dataset generation from videos.
\newblock In {\em ECCV}, 2022.

\bibitem{hassani2023neighborhood}
Ali Hassani, Steven Walton, Jiachen Li, Shen Li, and Humphrey Shi.
\newblock Neighborhood attention transformer.
\newblock In {\em CVPR}, 2023.

\bibitem{horn1981determining}
Berthold~KP Horn and Brian~G Schunck.
\newblock Determining optical flow.
\newblock {\em Artificial intelligence}, 1981.

\bibitem{huang2022flowformer}
Zhaoyang Huang, Xiaoyu Shi, Chao Zhang, Qiang Wang, Ka~Chun Cheung, Hongwei Qin, Jifeng Dai, and Hongsheng Li.
\newblock Flowformer: A transformer architecture for optical flow.
\newblock {\em arXiv:2203.16194}, 2022.

\bibitem{huang2019ccnet}
Zilong Huang, Xinggang Wang, Lichao Huang, Chang Huang, Yunchao Wei, and Wenyu Liu.
\newblock Ccnet: Criss-cross attention for semantic segmentation.
\newblock In {\em ICCV}, 2019.

\bibitem{hui2018liteflownet}
Tak-Wai Hui, Xiaoou Tang, and Chen~Change Loy.
\newblock Liteflownet: A lightweight convolutional neural network for optical flow estimation.
\newblock In {\em CVPR}, 2018.

\bibitem{hui2020lightweight}
Tak-Wai Hui, Xiaoou Tang, and Chen~Change Loy.
\newblock A lightweight optical flow cnn—revisiting data fidelity and regularization.
\newblock {\em TPAMI}, 2020.

\bibitem{Hur2019IterativeRR}
Junhwa Hur and S. Roth.
\newblock Iterative residual refinement for joint optical flow and occlusion estimation.
\newblock In {\em CVPR}, 2019.

\bibitem{jeong2022imposing}
Jisoo Jeong, Jamie~Menjay Lin, Fatih Porikli, and Nojun Kwak.
\newblock Imposing consistency for optical flow estimation.
\newblock In {\em CVPR}, 2022.

\bibitem{Jiang_2021_ICCV}
Shihao Jiang, Dylan Campbell, Yao Lu, Hongdong Li, and Richard Hartley.
\newblock Learning to estimate hidden motions with global motion aggregation.
\newblock In {\em ICCV}, 2021.

\bibitem{Jiang2021LearningOF}
Shihao Jiang, Yao Lu, Hongdong Li, and R. Hartley.
\newblock Learning optical flow from a few matches.
\newblock In {\em CVPR}, 2021.

\bibitem{Kondermann2016TheHB}
D. Kondermann, Rahul Nair, Katrin Honauer, Karsten Krispin, Jonas Andrulis, Alexander Brock, Burkhard G{\"u}ssefeld, Mohsen Rahimimoghaddam, Sabine Hofmann, C. Brenner, and B. J{\"a}hne.
\newblock The hci benchmark suite: Stereo and flow ground truth with uncertainties for urban autonomous driving.
\newblock In {\em CVPRW}, 2016.

\bibitem{li2023gyroflow}
Haipeng Li, Kunming Luo, Bing Zeng, and Shuaicheng Liu.
\newblock Gyroflow+: Gyroscope-guided unsupervised deep homography and optical flow learning.
\newblock {\em arXiv:2301.10018}, 2023.

\bibitem{liu2020arflow}
Liang Liu, Jiangning Zhang, Ruifei He, Yong Liu, Yabiao Wang, Ying Tai, Donghao Luo, Chengjie Wang, Jilin Li, and Feiyue Huang.
\newblock Learning by analogy: Reliable supervision from transformations for unsupervised optical flow estimation.
\newblock In {\em CVPR}, 2020.

\bibitem{liu2021asflow}
Shuaicheng Liu, Kunming Luo, Ao Luo, Chuan Wang, Fanman Meng, and Bing Zeng.
\newblock Asflow: Unsupervised optical flow learning with adaptive pyramid sampling.
\newblock {\em TCSVT}, 2021.

\bibitem{liu2021swin}
Ze Liu, Yutong Lin, Yue Cao, Han Hu, Yixuan Wei, Zheng Zhang, Stephen Lin, and Baining Guo.
\newblock Swin transformer: Hierarchical vision transformer using shifted windows.
\newblock In {\em ICCV}, 2021.

\bibitem{Luo_2022_CVPR}
Ao Luo, Fan Yang, Xin Li, and Shuaicheng Liu.
\newblock Learning optical flow with kernel patch attention.
\newblock In {\em CVPR}, 2022.

\bibitem{luo2022learning}
Ao Luo, Fan Yang, Kunming Luo, Xin Li, Haoqiang Fan, and Shuaicheng Liu.
\newblock Learning optical flow with adaptive graph reasoning.
\newblock In {\em AAAI}, 2022.

\bibitem{luo2021upflow}
Kunming Luo, Chuan Wang, Shuaicheng Liu, Haoqiang Fan, Jue Wang, and Jian Sun.
\newblock Upflow: Upsampling pyramid for unsupervised optical flow learning.
\newblock In {\em CVPR}, 2021.

\bibitem{luo2023learning}
Xinglong Luo, Kunming Luo, Ao Luo, Zhengning Wang, Ping Tan, and Shuaicheng Liu.
\newblock Learning optical flow from event camera with rendered dataset.
\newblock In {\em ICCV}, 2023.

\bibitem{Mayer2016ALD}
N. Mayer, Eddy Ilg, Philip H{\"a}usser, P. Fischer, D. Cremers, A. Dosovitskiy, and T. Brox.
\newblock A large dataset to train convolutional networks for disparity, optical flow, and scene flow estimation.
\newblock In {\em CVPR}, 2016.

\bibitem{KITTI_2015}
Moritz Menze and Andreas Geiger.
\newblock Object scene flow for autonomous vehicles.
\newblock In {\em CVPR}, 2015.

\bibitem{qin2022cosformer}
Zhen Qin, Weixuan Sun, Hui Deng, Dongxu Li, Yunshen Wei, Baohong Lv, Junjie Yan, Lingpeng Kong, and Yiran Zhong.
\newblock cosformer: Rethinking softmax in attention.
\newblock In {\em ICLR}, 2022.

\bibitem{richter2017playing}
Stephan~R Richter, Zeeshan Hayder, and Vladlen Koltun.
\newblock Playing for benchmarks.
\newblock In {\em ICCV}, 2017.

\bibitem{sui2022craft}
Xiuchao Sui, Shaohua Li, Xue Geng, Yan Wu, Xinxing Xu, Yong Liu, Rick Goh, and Hongyuan Zhu.
\newblock Craft: Cross-attentional flow transformer for robust optical flow.
\newblock In {\em CVPR}, 2022.

\bibitem{sun2022disentangling}
Deqing Sun, Charles Herrmann, Fitsum Reda, Michael Rubinstein, David~J Fleet, and William~T Freeman.
\newblock Disentangling architecture and training for optical flow.
\newblock In {\em ECCV}, 2022.

\bibitem{sun2021autoflow}
Deqing Sun, Daniel Vlasic, Charles Herrmann, Varun Jampani, Michael Krainin, Huiwen Chang, Ramin Zabih, William~T Freeman, and Ce Liu.
\newblock Autoflow: Learning a better training set for optical flow.
\newblock In {\em CVPR}, 2021.

\bibitem{Sun2018PWCNetCF}
Deqing Sun, Xiaodong Yang, Ming-Yu Liu, and J. Kautz.
\newblock Pwc-net: Cnns for optical flow using pyramid, warping, and cost volume.
\newblock In {\em CVPR}, 2018.

\bibitem{sun2022skflow}
Shangkun Sun, Yuanqi Chen, Yu Zhu, Guodong Guo, and Ge Li.
\newblock Skflow: Learning optical flow with super kernels.
\newblock In {\em NeurIPS}, 2022.

\bibitem{Teed2020RAFTRA}
Zachary Teed and Jun Deng.
\newblock Raft: Recurrent all-pairs field transforms for optical flow.
\newblock In {\em ECCV}, 2020.

\bibitem{Wang2020DisplacementInvariantMC}
Jianyuan Wang, Yiran Zhong, Yuchao Dai, K. Zhang, Pan Ji, and Hongdong Li.
\newblock Displacement-invariant matching cost learning for accurate optical flow estimation.
\newblock In {\em NeurIPS}, 2020.

\bibitem{wang2018non}
Xiaolong Wang, Ross Girshick, Abhinav Gupta, and Kaiming He.
\newblock Non-local neural networks.
\newblock In {\em CVPR}, 2018.

\bibitem{wang2020non}
Zhengyang Wang, Na Zou, Dinggang Shen, and Shuiwang Ji.
\newblock Non-local u-nets for biomedical image segmentation.
\newblock In {\em AAAI}, 2020.

\bibitem{weinzaepfel2013deepflow}
Philippe Weinzaepfel, Jerome Revaud, Zaid Harchaoui, and Cordelia Schmid.
\newblock Deepflow: Large displacement optical flow with deep matching.
\newblock In {\em ICCV}, 2013.

\bibitem{witkin1984scale}
Andrew Witkin.
\newblock Scale-space filtering: A new approach to multi-scale description.
\newblock In {\em ICASSP}, 1984.

\bibitem{xu2021high}
Haofei Xu, Jiaolong Yang, Jianfei Cai, Juyong Zhang, and Xin Tong.
\newblock High-resolution optical flow from 1d attention and correlation.
\newblock In {\em ICCV}, 2021.

\bibitem{xu2022gmflow}
Haofei Xu, Jing Zhang, Jianfei Cai, Hamid Rezatofighi, and Dacheng Tao.
\newblock Gmflow: Learning optical flow via global matching.
\newblock In {\em CVPR}, 2022.

\bibitem{Yang2019VolumetricCN}
Gengshan Yang and D. Ramanan.
\newblock Volumetric correspondence networks for optical flow.
\newblock In {\em NeurIPS}, 2019.

\bibitem{yu2016back}
Jason~J Yu, Adam~W Harley, and Konstantinos~G Derpanis.
\newblock Back to basics: Unsupervised learning of optical flow via brightness constancy and motion smoothness.
\newblock In {\em ECCV}, 2016.

\bibitem{zhang2021separable}
Feihu Zhang, Oliver~J Woodford, Victor~Adrian Prisacariu, and Philip~HS Torr.
\newblock Separable flow: Learning motion cost volumes for optical flow estimation.
\newblock In {\em ICCV}, 2021.

\bibitem{zhao2020msrn}
Chengqian Zhao, Cheng Feng, Dengwang Li, and Shuo Li.
\newblock Of-msrn: optical flow-auxiliary multi-task regression network for direct quantitative measurement, segmentation and motion estimation.
\newblock In {\em AAAI}, 2020.

\bibitem{Zhao2020MaskFlownetAF}
Shengyu Zhao, Yilun Sheng, Yue Dong, E. Chang, and Yan Xu.
\newblock Maskflownet: Asymmetric feature matching with learnable occlusion mask.
\newblock In {\em CVPR}, 2020.

\bibitem{zhao2022global}
Shiyu Zhao, Long Zhao, Zhixing Zhang, Enyu Zhou, and Dimitris Metaxas.
\newblock Global matching with overlapping attention for optical flow estimation.
\newblock In {\em CVPR}, 2022.

\bibitem{zheng2022dip}
Zihua Zheng, Ni Nie, Zhi Ling, Pengfei Xiong, Jiangyu Liu, Hao Wang, and Jiankun Li.
\newblock Dip: Deep inverse patchmatch for high-resolution optical flow.
\newblock In {\em CVPR}, 2022.

\bibitem{zhu2019asymmetric}
Zhen Zhu, Mengde Xu, Song Bai, Tengteng Huang, and Xiang Bai.
\newblock Asymmetric non-local neural networks for semantic segmentation.
\newblock In {\em CVPR}, 2019.

\end{thebibliography}
}

\end{document}